\documentclass[10pt,twocolumn,letterpaper]{article}

\usepackage{cvpr}
\usepackage{times}
\usepackage{epsfig}
\usepackage{graphicx}
\usepackage{amsmath}
\usepackage{amssymb}

\usepackage{algorithm}
\usepackage[noend]{algpseudocode}
\usepackage{booktabs}
\usepackage{multirow}
\usepackage{makecell}
\usepackage{xparse}
\usepackage{fancyhdr}
\usepackage[pagebackref=true,breaklinks=true,letterpaper=true,colorlinks,bookmarks=false]{hyperref}

\newcommand{\argmin}[1]{\underset{#1}{\operatorname{arg}\,\operatorname{min}}\;}
\newcommand{\argmax}[1]{\underset{#1}{\operatorname{arg}\,\operatorname{max}}\;}
\newcommand{\norm}[1]{\left\lVert#1\right\rVert}
\NewDocumentCommand{\rot}{O{45} O{1em} m}{\makebox[#2][l]{\rotatebox{#1}{#3}}}%

\cvprfinalcopy % *** Uncomment this line for the final submission

\def\cvprPaperID{3780} % *** Enter the CVPR Paper ID here

\ifcvprfinal\fi
\begin{document}

\title{A Weighted Sparse Sampling and Smoothing Frame Transition Approach for Semantic Fast-Forward First-Person Videos}

\author{Michel Silva \and Washington Ramos \and Joao Ferreira \and Felipe Chamone \and Mario Campos \and Erickson R. Nascimento \\
	Universidade Federal de Minas Gerais (UFMG), Brazil\\
	{\tt\small \{michelms, washington.ramos, joaoklock, cadar, mario, erickson\}@dcc.ufmg.br}
}
\maketitle

\thispagestyle{fancy}
\fancyhf{}
\chead{{In Proceedings of the IEEE Conference on Computer Vision and Pattern Recognition (CVPR) 2018 \\ The final publication is available at: \href{https://doi.org/10.1109/CVPR.2018.00253}{doi.org/10.1109/CVPR.2018.00253}}}

%%%%%%%%% ABSTRACT
\begin{abstract}

Thanks to the advances in the technology of low-cost digital cameras and the popularity of the self-recording culture, the amount of visual data on the Internet is going to the opposite side of the available time and patience of the users. Thus, most of the uploaded videos are doomed to be forgotten and unwatched in a computer folder or website. In this work, we address the problem of creating smooth fast-forward videos without losing the relevant content. We present a new adaptive frame selection formulated as a weighted minimum reconstruction problem, which combined with a smoothing frame transition method accelerates first-person videos emphasizing the relevant segments and avoids visual discontinuities. The experiments show that our method is able to fast-forward videos to retain as much relevant information and smoothness as the state-of-the-art techniques in less time. We also present a new $80$-hour multimodal (RGB-D, IMU, and GPS) dataset of first-person videos with annotations for recorder profile, frame scene, activities, interaction, and attention\footnote{\href{https://www.verlab.dcc.ufmg.br/semantic-hyperlapse/cvpr2018/}{https://www.verlab.dcc.ufmg.br/semantic-hyperlapse/cvpr2018/}}.

\end{abstract}

\section{Introduction}
\label{sec:introduction}

By 2019, the online video might be responsible for more than 80\% of global Internet traffic~\cite{CISCO2016}. Not only are internet users watching more online video, but they are also recording themselves and producing a growing number of videos for sharing their day-to-day life routine. The ubiquity of inexpensive shoot video devices and the lower costs of producing and storing videos are giving unprecedented freedom to the people to create increasingly long-running first-person videos. On the other hand, such freedom might lead the user to create a final long-running and boring video, once most everyday activities do not merit recording.

A central challenge is to selective highlight the meaningful parts of the videos without losing the whole message that the video should convey. Although video summarization techniques~\cite{Molino2017, Mahasseni2017} provide quick access to videos' information, they only return segmented clips or single images of the relevant moments. By not including the very last and the following frames of a clip, the summarization might lose the clip context~\cite{Plummer2017}. Hyperlapse techniques yield quick access to the meaningful parts and also preserve the whole video context by fast-forwarding the videos applying an adaptive frame selection~\cite{Kopf2014,Joshi2015,Poleg2015}. Despite the Hyperlapse techniques being able to address the shake effects of fast-forwarding first-person videos, handling every frame equally important is a major weakness of these techniques. In a lengthy stream recorded using the always-on mode, some portions of the videos are undoubtedly more relevant than others. 
%%%
\begin{figure}[!t]
	\centering
	\includegraphics[width=1.0\linewidth]{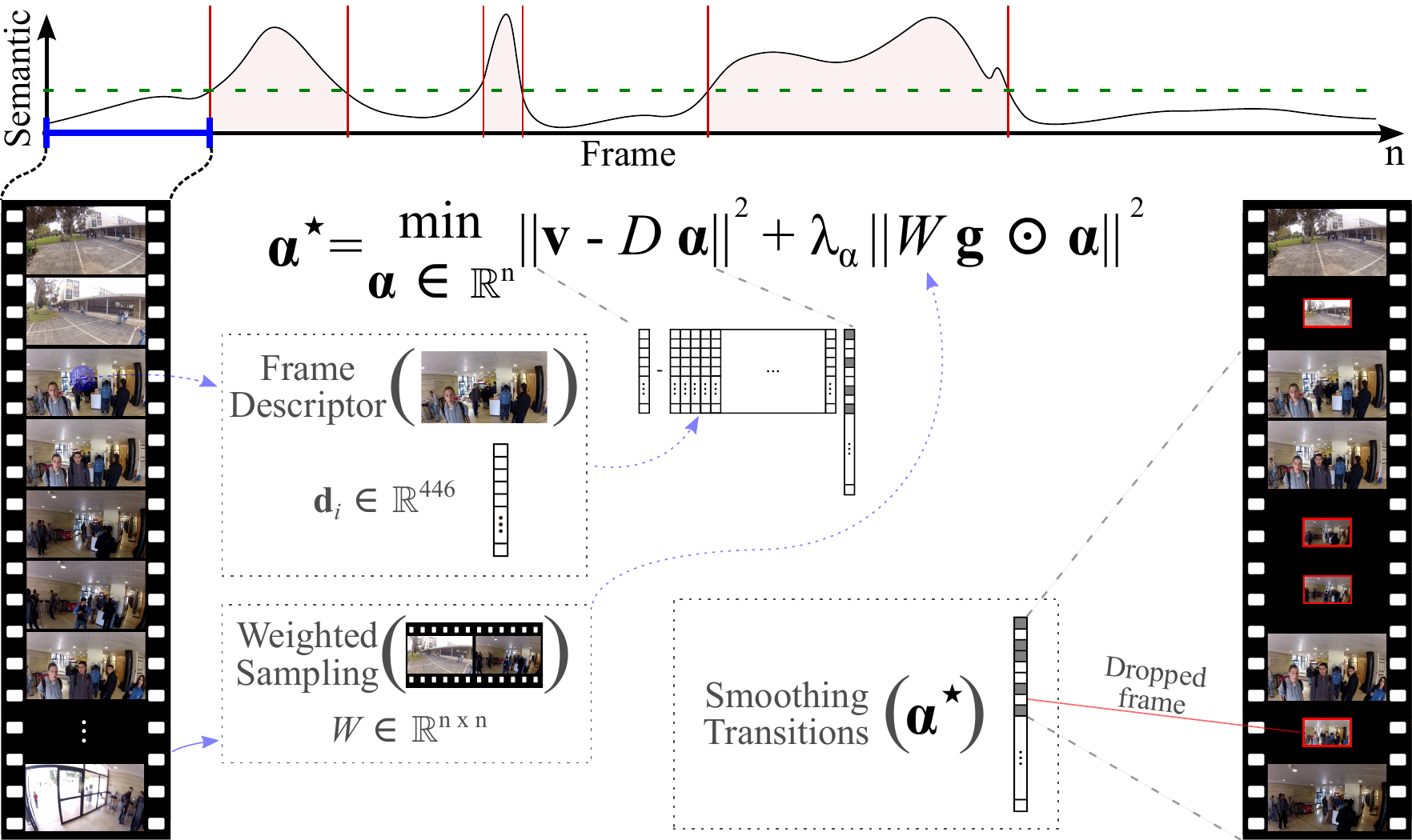}
	\caption{The fast-forward methodology. A weighted sampling combined with a smoothing transition step is applied to tackle the abrupt camera movements. The activation vector indicates which frames compose the fast-forward video. A smoothing step is applied to the transitions between the selected frames.}
	\label{fig:introduction}
\end{figure}
%%%

Most recently, methods on fast-forward videos emphasizing relevant content have emerged as promising and effective approaches to deal with the tasks of visual smoothness and semantic highlighting of first-person videos. The relevant information is emphasized by playing faster the non-semantic segments and applying a smaller speed-up rate in the semantic ones~\cite{Ramos2016,Silva2016,Lai2017,Silva2018} or even playing them in slow-motion~\cite{Yao2016}. 
To reach both objectives, visual smoothness and semantic highlight, these techniques describe the video frames and their transitions by features, and then formulate an optimization problem using the combination of these features. Consequently, the computation time and memory usage are impacted by the number of features used, once the search space grows exponentially. Therefore, the current Hyperlapse methods are not scalable regarding the number of features.

In this work, we present a new semantic fast-forward method that solves the adaptive frame sampling by modeling the frame selection as a Minimum Sparse Reconstruction (MSR) problem (Figure~\ref{fig:introduction}). The video is represented as a dictionary, where each column describes a video frame. The frames selection is defined by the activation vector, and the fast-forwarding effect is reached by the sparsity nature of the problem. In other words, we look for the smallest set of frames that provides the reconstruction of the original video with small error. Additionally, to attenuate abrupt camera movements in the final video, we apply a weighted version of the MSR problem, where frames related to camera movement are more likely to be sampled. 

In the proposed modeling, the scalability of features is not a problem anymore, because using a high dimensional descriptor leads to a balance of the dictionary dimensions, which is recommended to solve the MSR problem, and do not substantially affect the computational cost and memory usage. We experimentally demonstrate that our approach creates videos composed of more relevant information than the state-of-the-art Semantic Fast-Forwarding method and as smooth as the state-of-the-art Hyperlapse techniques.

The contributions of our work are: $i$) a set of methods capable of handling larger feature vectors to better describe the frames and the video transitions, addressing the abrupt camera motions while not increasing the computational processing time; $ii$) a new labeled 80-hour multimodal (3D Inertial Movement Unit, GPS, and RGB-D camera) dataset of first-person videos covering a wide range of activities such as video actions, party, beach, tourism, and academic life. Each frame is labeled with respect to the activity, scene, recorder ID, interaction, and attention.

\section{Related Work}
\label{sec:related_work}

Works on selective highlighting of the meaningful parts of first-person videos have been extensively studied in the past few years. We can broadly classify them into Video Summarization and Hyperlapse approaches.

\paragraph{Video Summarization.} 
The goal of video summarization is to produce a compact visual summary containing the most discriminative and informative parts of the original video. Techniques typically use features that range from low-level such as motion and color~\cite{Zhao2014, Gygli2015} to high-level (\eg, important objects, user preferences)~\cite{Kim2014, Yao2016, Sharghi2017}. 
Lee~\etal~\cite{Lee2012}~exploit interaction level, gaze, and object detection frequency as egocentric properties to create a storyboard of keyframes with important people and objects. Lu~and~Grauman~\cite{Lu2013}~present video skims as summaries instead of static keyframes. After splitting the video into subshots, they compute the mutual influence of objects and estimate the subshots importance to select the optimal chain of subshots.

Recent approaches are based on highlight detection~\cite{Lin2015, Bettadapura2016, Yao2016} and vision-language models~\cite{Sharghi2017, Plummer2017, Panda2017}. Bettadapura~\etal~\cite{Bettadapura2016}~propose an approach for identifying picturesque highlights. They use composition, symmetry and color vibrancy as scoring metrics and leverage GPS data to filter frames by the popularity of the location. Plummer~\etal~\cite{Plummer2017}~present a semantically-aware video summarization. They optimize a linear combination of visual, \ie, representativeness,  uniformity, interestingness, and vision-language objectives to select the best subset of video segments.

Sparse Coding has been successfully applied to many varieties of vision tasks~\cite{Wright2009, Zhao2011, Cong2012, Zhao2014, oliveira_tip2014, Mei2014, Mei2015pr}. In video summarization, Cong~\etal~\cite{Cong2012}~formulate the problem of video summarization as a dictionary selection problem. They propose a novel model to either extract keyframes or generate video skims using sparsity consistency. Zhao~\etal~\cite{Zhao2014}~propose a method based on online dictionary learning that generates summaries on-the-fly. They use sparse coding to eliminate repetitive events and create a representative short version of the original video. The main benefit of using sparse coding for frame selection is that selecting a different number of frames does not incur an additional computational cost. This work differs from sparse coding video summarization since it handles the shakiness in the transitions via a weighted sparse frame sampling solution. Also, it is capable of dealing with the temporal gap caused by discontinuous skims. 

\paragraph{Hyperlapse.} 
A pioneering work in creating hyperlapse from casual first-person videos was conducted by Kopf~\etal~\cite{Kopf2014}. The output video comes from the use of image-based rendering techniques such as projecting, stitching and blending after computing the optimal trajectory of the camera poses. Despite their remarkable results, the method has a high computational cost and requires camera motion and parallax to compute the 3D model of the scene.

Recent strategies focus on selecting frames using different adaptive approaches to adjust the density of frame selection according to the cognitive load. Poleg~\etal~\cite{Poleg2015}~model the frame selection as a shortest path in a graph. The nodes of this graph represent the frames of the original video and, the edges weights between pairs of frames are proportional to the cost of including the pair sequentially in the output video. An extension for creating a panoramic hyperlapse of a single or multiple input videos was proposed by Halperin~\etal~\cite{Halperin2017}. They enlarge each of the input frames using neighboring frames from the videos to reduce the perception of instability. Joshi~\etal~\cite{Joshi2015} present a method based on dynamic programming to select an optimal set of frames regarding the desired target speed-up and the smoothness in frame-to-frame transitions jointly. 

Although these solutions have succeeded in creating short and watchable versions of long first-person videos, they often remove segments of high relevance to the user, since the methods handle all frames as having the same semantic relevance.

\paragraph{Semantic Hyperlapse.} 
Unlike traditional hyperlapse techniques, where the goal is to optimize the output number of frames and the visual smoothness, the semantic hyperlapse techniques also include the semantic relevance for each frame. Ramos~\etal~\cite{Ramos2016}~introduced an adaptive frame sampling process embedding semantic information within. The methodology assigns scores to frames based on the detection of predefined objects that may be relevant to the recorder. The rate of dropped frames is a function of the relative semantic load and the visual smoothness. Later, Silva~\etal~\cite{Silva2016} extended the Ramos~\etal's method using a better semantic temporal segmentation and an egocentric video stabilization process in the fast-forward output. The drawbacks of these works include abrupt changes in the acceleration and shaky exhibition at every large lateral swing in the camera.

Most recently, two new hyperlapse methods for first-person videos were proposed: the Lai~\etal's system~\cite{Lai2017} and the Multi-Importance Fast-Forward (MIFF)~\cite{Silva2018} method. Lai~\etal's system converts $ 360^\circ $ videos into normal field-of-view hyperlapse videos. They extract semantics through regions of interest using spatial-temporal saliency and semantic segmentation to guide camera path planning. Low rates of acceleration are assigned to interesting regions to emphasize them in the hyperlapse output. In the MIFF method, the authors applied a learning approach to infer the users' preference and determine the relevance of a given frame. The MIFF calculates different speed-up rates for segments of the video, which are extracted using an iterative temporal segmentation process according to the semantic content.

Although not focused on the creation of hyperlapses, Yao~\etal~\cite{Yao2016}~present a highlight-driven summarization approach that generates skimming and timelapse videos as summaries from first-person videos. They assign scores to the video segments by using late fusion of spatial and temporal deep convolution neural networks (DCNNs). The segments with higher scores are selected as video highlights. For the video timelapse, they calculate proper speedup rates such that the summary is compressed in the non-highlight segments and expanded in highlight segments. It is noteworthy that timelapse videos do not handle the suavity constraint that is a mandatory requirement for hyperlapse videos. Differently from the aforementioned work, our approach optimizes semantic, length and smoothness to create semantic hyperlapses. Most importantly, it keeps the path taken by the recorder avoiding to lose the flow of the story and thus, conveying the full message from the original video in a shorter and smoother version.

\section{Methodology}
\label{sec:methodology}

In this section, we describe a new method for creating smooth fast-forward videos that retains most of the semantic content of the original video in a reduced processing time. Our method consists of four primary steps: i) Creation and temporal segmentation of a semantic profile of the input video; ii) Weighted sparse frame sampling; iii) Smoothing Frame Transitions (SFT), and iv) Video compositing. 
\begin{figure*}
	\centering
	\includegraphics[width=0.975\linewidth]{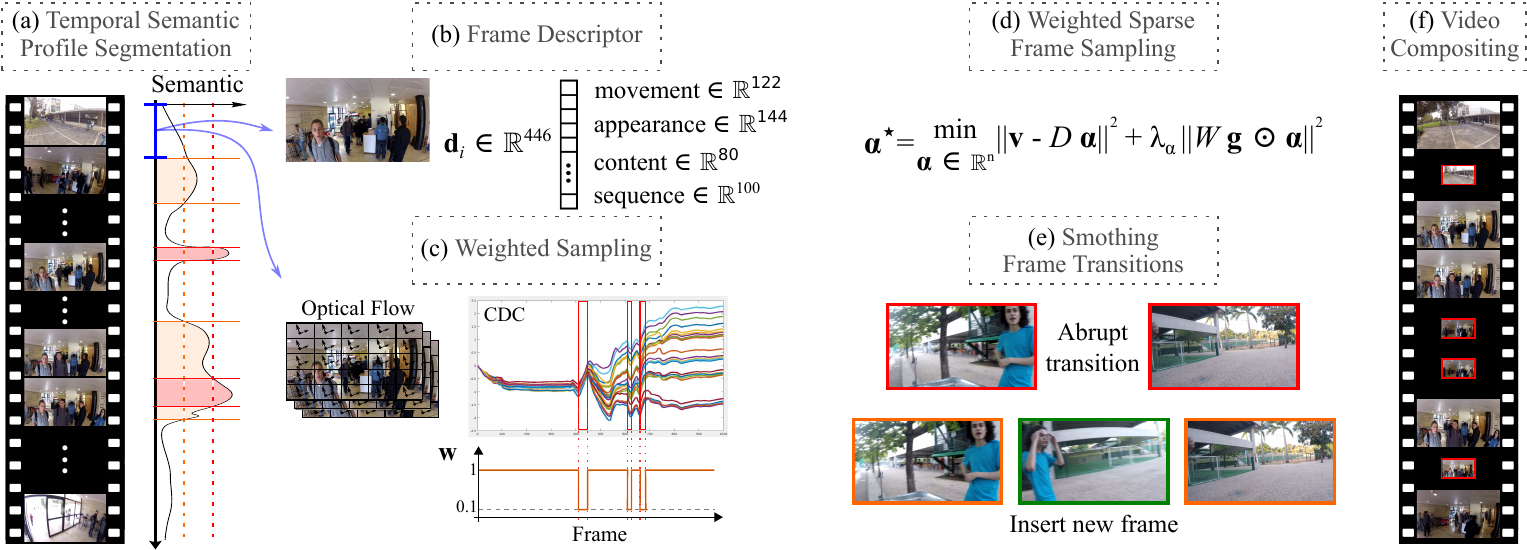}
	\caption{Main steps of our fast-forward methodology. For each segment created in the temporal semantic profile segmentation (a), the frames are described (b) and weighted based on the camera movement computed (c). The frames are sampled by minimizing local-constrained and reconstruction problem (d). The smoothing step is applied to tackle the abrupt transitions of the selected frames (e).}
	\label{fig:methodology}
\end{figure*}
\subsection{Temporal Semantic Profile Segmentation}

The first step of a semantic fast-forward method is the creation of a semantic profile of the input video. Once we set a semantic (\eg, faces, type of objects of interest, scene, \etc), a video score profile is created by extracting the relevant information and assigning a semantic score for each frame of the video (Figure~\ref{fig:methodology}-a). The confidence of the classifier combined with the locality and size of the regions of interest score are used as the semantic score~\cite{Ramos2016,Silva2016}.

The set of scores defines a profile curve, which is used for segmenting the input video into semantic and non-semantic sequences. Following, a refinement process is executed in the semantic segments, creating levels of importance regarding the defined semantic. Finally, speed-up rates are calculated based on the length and level of relevance of each segment. The rates are calculated such that the semantic segments are played slower than the non-semantic ones, and the whole video achieves the desired speed-up. We refer the reader to~\cite{Silva2018} for a more detailed description of the multi-importance semantic segmentation.

The output of this step is a set of segments that are used to feed the following steps that process each one separately.

\subsection{Weighted Sparse Frame Sampling} 

In general, hyperlapse techniques solve the adaptive frame selection problem searching the optimal configuration (\eg, shortest path in a graph or dynamic programming) in a space of representation where different types of features are combined to represent frames or transitions between frames. A large number of features can be used for improving the representation of a frame or transitions, but such solution leads to a high-dimensional representation space increasing the computation time and memory usage. We address this problem of representation using a sparse frame sampling approach, Figure~\ref{fig:methodology}-d.

Let ${D=[\mathbf{d}_1, \mathbf{d}_2, \mathbf{d}_3, \cdots, \mathbf{d}_n] \in \mathbb{R}^{f \times n}}$ be a segment of the original video with $n$ frames represented in our feature space. Each entry ${\mathbf{d}_i \in \mathbb{R}^{f}}$ stands for the feature vector of the ${i}$-th frame. Let the video story ${\mathbf{v} \in \mathbb{R}^{f}}$ be defined as the sum of the frame features of the whole segment, \ie, ${\mathbf{v} = \sum_{i=1}^n \mathbf{d}_i}$. The goal is to find an optimal subset ${S=[\mathbf{d}_{s_1},\mathbf{d}_{s_2}, \mathbf{d}_{s_3},\cdots,\mathbf{d}_{s_m}] \in \mathbb{R}^{f \times m}}$, where ${m \ll n}$ and ${\{s_1,s_2,s_3,\cdots,s_m\}}$ belongs to the set of frames in the segment. 

Let the vector~${\boldsymbol{\alpha} \in \mathbb{R}^{n}}$ be an activation vector indicating whether ${\mathbf{d}_i}$ is in the set $S$ or not. The problem of finding the values for $\boldsymbol{\alpha}$ that lead to a small reconstruction error of $\mathbf{v}$, can be formulated as a Locality-constrained Linear Coding (LLC)~\cite{Wang2010} problem as follow:
\begin{equation}
	\label{eq:LLC}
	\argmin{\boldsymbol{\alpha}~\in~\mathbb{R}^{n}} { \norm{\mathbf{v} - D~\boldsymbol{\alpha}}^{2} + \lambda_\alpha \norm{\mathbf{g} \odot \boldsymbol{\alpha}}^2 },
\end{equation}
where $\mathbf{g}$ is the Euclidean distance of each dictionary entry $\mathbf{d}_i$ to the segment representation $\mathbf{v}$, and $\odot$ is an element-wise multiplication operator. The ${\lambda_{\alpha}}$ is the regularization term of the locality of the vector $\boldsymbol{\alpha}$. 

The benefit of using LLC formulation instead of the traditional Sparse Coding (SC) model is twofold. The LLC provides local smooth sparsity and can be solved by an analytical solution, which results in a smoother final fast-forward video in a lower computational cost. 

\paragraph{Weighted Sampling.} 
Abrupt camera motions are challenging issues for fast-forwarding video techniques. They might lead to the creation of shaky  and nauseating videos. To tackle this issue, we used a weighted Locality-constrained Linear Coding formulation, where each dictionary entry has a weight assigned to it:
\begin{equation}
	\label{eq:LLCW}
	\boldsymbol{\alpha^\star} = \argmin{\boldsymbol{\alpha}~\in~\mathbb{R}^{n}} { \norm{\mathbf{v} - D~\boldsymbol{\alpha}}^{2} + \lambda_\alpha \norm{W~\mathbf{g} \odot \alpha}^2 },
\end{equation}
where $W$ is a diagonal matrix built from the weight vector ${\mathbf{w} \in \mathbb{R}^n}$, \ie, ${W\triangleq\text{diag}(\mathbf{w})}$.

This weighting formulation provides a flexible solution, where we create different weights for frames based on the camera movement and thus, we can change the contribution for the reconstruction without increasing the sparsity/locality term significantly. 

Let ${C \in \mathbb{R}^{c \times n}}$ be the Cumulative Displacement Curves~\cite{Poleg2014}, \ie, the cumulative sum of the Optical Flow magnitudes, computed over the horizontal displacements in ${5\times5}$ grid windows of the video frames (Figure~\ref{fig:methodology}-c). Let ${C' \in \mathbb{R}^{c \times n}}$ be the derivative of each curve $C$ \wrt time. We assume frame $i$ to be within an interval of abrupt camera motion if all curves $C'$ present the same sign (positive/negative) at the point $i$, which represents a head-turning movement~\cite{Poleg2014}. We assign a lower weight for these motion intervals to enforce them to be composed of a larger number of frames. We empirically set the weights to ${\mathbf{w}_i=0.1}$ and ${\mathbf{w}_i=1.0}$ for the frame features inside and outside the interval, respectively.

\paragraph{Speed-up Selection.}
All frames related to the activated positions of the vector~$\boldsymbol{\alpha^\star}$ will be selected to compose the final video. Since ${\lambda_\alpha}$ controls the sparsity, it also manages the speed-up rate of the created video. The zero-value ${\lambda_\alpha}$ enables the activation of all frames leading to a complete reconstruction. 
To achieve the desired speed-up, we perform an iterative search starting from zero, as depicted in Algorithm~\ref{alg:lambda_adjustment}. The function ${NumOfFrames(\lambda)}$ (Line~\ref{alg_line:number_selected_frames}) solves Equation~\ref{eq:LLCW} using $\lambda$ as the value of $\lambda_\alpha$ and returns the number of activations in $\boldsymbol{\alpha^\star}$.
\begin{algorithm}[!t]
	\caption{Lambda value adjustment}
	\label{alg:lambda_adjustment}
	\begin{algorithmic}[1]
		\Require \small{Desired length of the final video ${VideoLength}$.}
		\Ensure \small{The ${\lambda_\alpha}$ value to reach the desired number of frames.}
		\Function{Lambda\_Adjustment}{${VideoLength}$}
		\State ${\lambda_\alpha \gets 0}$ , ${step \gets 0.1}$ , ${nFrames \gets 0}$
		%		\While{${nf \neq Sd} \lor m > 10^{-15}$}
		\While{${nFrames \neq VideoLength} $}
		\State $ nFrames \gets NumberOfFrames(\lambda_\alpha + step)$
		\label{alg_line:number_selected_frames}
		\If {${nFrames \geq VideoLength}$}
		\State ${\lambda_\alpha \gets \lambda_\alpha + step}$
		\Else
		\State ${step \gets step / 10}$
		\EndIf
		\EndWhile
		\EndFunction
	\end{algorithmic}
\end{algorithm}
\paragraph{Frame Description.}
The feature vector of the $i$-th frame ${\mathbf{d}_i \in \mathbb{R}^{446}}$ (Figure~\ref{fig:methodology}-b) is composed of the concatenation the following terms. 
The ${\mathbf{hof_m} \in \mathbb{R}^{50}}$ and ${\mathbf{hof_o} \in \mathbb{R}^{72}}$ are histogram of optical flow magnitudes and orientations of the $i$-th frame, respectively. The appearance descriptor, ${\mathbf{a} \in \mathbb{R}^{144}}$, is composed of the mean, standard deviation, and skewness values of HSV color channels of the windows in a ${4\times4}$ grid of the frame $i$. 
To define the content descriptor, ${\mathbf{c} \in \mathbb{R}^{80}}$, we first use the YOLO~\cite{Redmon2016} to detect the objects in the frame $i$; then, we create a histogram with these objects over the 80 classes of the YOLO architecture.
Finally, the sequence descriptor, ${\mathbf{s} \in \mathbb{R}^{100}}$, is an one hot vector, with the ${\text{mod}(i,100)}$-th feature activated. 

\subsection{Smoothing Frame Transitions}

A solution ${\boldsymbol{\alpha^\star}}$ does not ensure a final smooth fast-forward video. Occasionally, the solution might provide a low error reconstruction of small and highly detailed segments of the video. Thus, by creating a better reconstruction with a limited number of frames, ${\boldsymbol{\alpha^\star}}$ might ignore stationary moments or visually likely views and create videos similar to results of summarization methods.

We address this issue by dividing the frame sampling into two steps. First, we run the weighted sparse sampling to reconstruct the video using a speed-up multiplied by a factor ${SpF}$. The resulting video contains ${1/SpF}$ of the desired number frames. Then, we iteratively insert frames into the shakier transitions (Figure~\ref{fig:methodology}-e) until the video achieves the exact number of frames.

Let ${I(F_x,F_y)}$ be the instability function defined by ${I(F_x,F_y)= AC(F_x,F_y)*(d_{y}-d_{x}-speedup)}$. The function ${AC(F_x, F_y)}$ calculates the Earth Mover's Distance~\cite{Pele2009} between the color histograms of the frames $F_x$ and $F_y$. The second term of the instability function is the speed-up deviation term. This term calculates how far the distance between frames $F_x$ and $F_y$, \ie, ${d_y - d_x}$, are from the desired speedup. We identify a shakier transition using the Equation~\ref{eq:identify_peak}:
\begin{equation}
	\label{eq:identify_peak}
	{i^\star = \argmax{i~\in~\mathbb{R}^{m}}{I(F_{s_i},F_{s_{i+1}})}}.
	%	{i^\star = \argmax{i~\in~\mathbb{R}^{m} }{AC(F_{d_{si}},F_{d_{si+1}}) * sd } },
\end{equation}
The set of frames from ${F_{s_{i^\star}}}$~to~${F_{s_{i^\star+1}}}$, \ie, solution of Equation~\ref{eq:identify_peak}, has visually dissimilar frames with a distance between them higher than the required speed-up. 

After identifying the shakier transition, from the subset with frames ranging from ${F_{s_{i^\star}}}$ to ${F_{s_{i^\star+1}}}$, we choose the frame ${F_{j^\star}}$ that minimizes the instability of the frame transition as follows: 
\begin{equation}
	\label{eq:frame_picker}
	{j^\star = \argmin{j~\in~\mathbb{R}^{n}}{I(F_{s_{i^\star}},F_j)^2 + I(F_j,F_{s_{i^\star+1}})^2} }.
\end{equation}
Equations~\ref{eq:identify_peak}~and~\ref{eq:frame_picker} can be solved by exhaustive search, since the interval is small. In this work, we use ${SpF=2}$ in the experiments. Higher values enlarge the search interval, increasing the time for solving Equation~\ref{eq:frame_picker}.

\subsection{Video compositing}

All selected frames of each segment are concatenated to compose the final video (Figure~\ref{fig:methodology}-f). In this last step, we also run the egocentric video stabilization proposed in the work of Silva~\etal~\cite{Silva2016}, which is properly designed to fast-forwarded egocentric videos. The stabilizer creates smooth transitions by applying weighted homographies. Images corrupted during the smoothing step are reconstructed using the non-selected frames of the original video.

\section{Experiments}
\label{sec:exp_results}

In this section, we describe the experimental results on the Semantic Dataset~\cite{Silva2016} and a new multimodal semantic egocentric dataset. After detailing the datasets, we present the results followed by the ablation study on the components and efficiency analysis.
\begin{figure}[!t]
	\centering
	\includegraphics[width=1.0\linewidth]{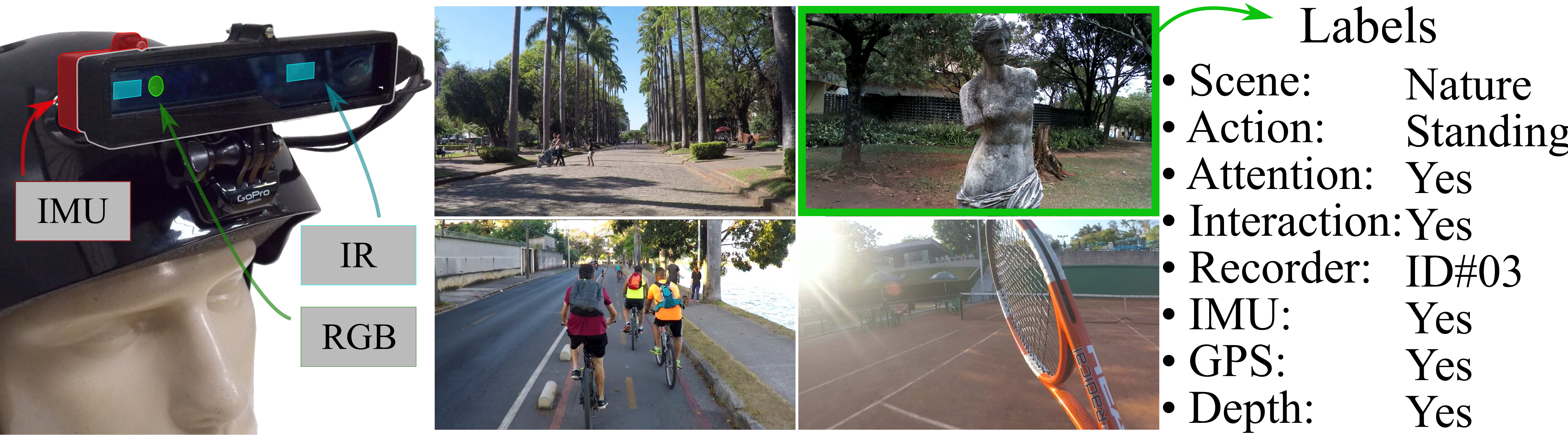}
	\caption{Left: setup used to record videos with RGB-D camera and IMU. Center: frame samples from DoMSEV. Right: an example of the available labels for the image highlighted in green.}
	\label{fig:dataset}
\end{figure}
\begin{figure*}[!t]
	\centering
	\includegraphics[width=0.883\linewidth]{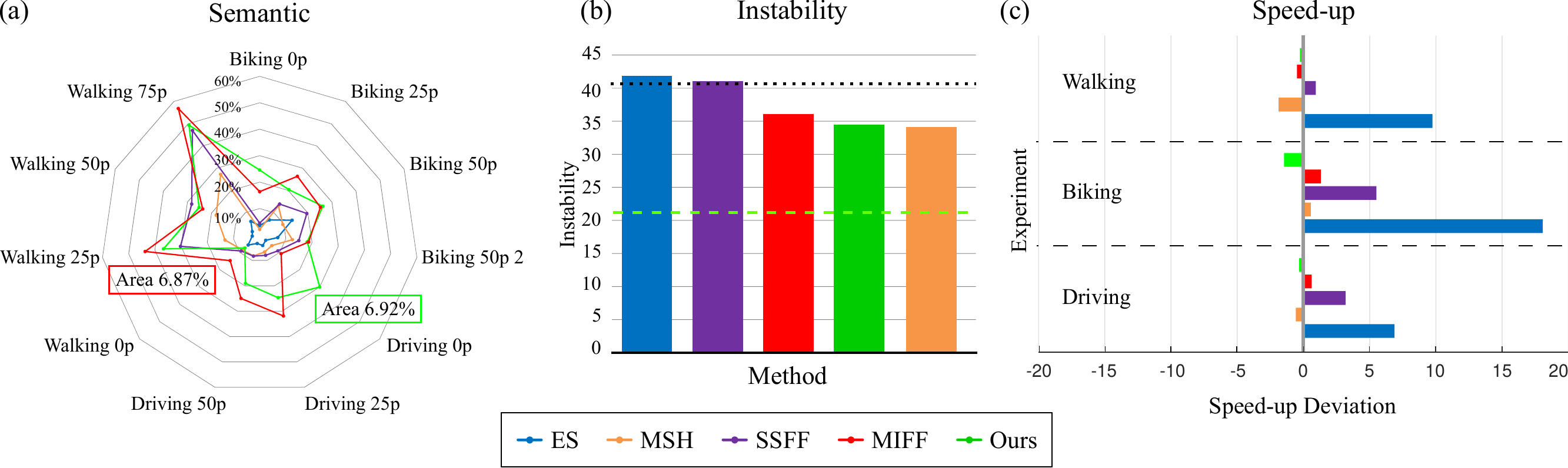}
	\caption{Evaluation of the proposed Sparse Sampling methodology against the literature fast-forward methods. Dashed and doted lines in (b) are related to the mean instability indexes of the original video and the uniform sampling, respectively. Desirable values are: (a) higher, (b) lower, and (c) closer to zero.}
	\label{fig:results_baselines}
\end{figure*}
\subsection{Datasets and Evaluation criterion}

\paragraph{Semantic Dataset.} 
We first test our method using the Semantic Dataset, proposed by Silva~\etal~\cite{Silva2016}. This dataset is composed of $11$ annotated videos. Each video is classified having  $0\%$, $25\%$, $50\%$, or $75\%$ of semantic content in the semantic portions (a set of frames with high semantic score) on average. For instance, in the Walking25p video, the recorder is walking and there are an average of $25\%$ of frames with faces and/or pedestrians. It is worth noting that even when video belongs to the class 0p, it still contains semantics on its frames. The reason of being classified as $0$p is mainly because it does not have a minimum number of frames with high semantic score. 

Because this dataset has the annotation of the semantic load, we can use it for finding the best semantic fast-forward method, \ie, the fast-forward approach that retains the highest semantic load of the original video.

\paragraph{Multimodal Semantic Egocentric Videos.} 
Aside from the Semantic Dataset, we also evaluated our approach on new 80-hour data set. Because of the absence of unrestricted and available multimodal data to work with egocentric tasks, we propose an 80-hour Dataset of Multimodal Semantic Egocentric Videos (DoMSEV). The videos cover a wide range of activities such as shopping, recreation, daily life, attractions, party, beach, tourism, sports, entertainment, and academic life. 

The multimodal data was recorded using either a GoPro Hero\textsuperscript\texttrademark camera or a built setup composed of a 3D Inertial Movement Unit (IMU) attached to the Intel Realsense\textsuperscript\texttrademark R200 RGB-D camera. Figure~\ref{fig:dataset} shows the setup used and a few of examples of frames from the videos. Different people recorded the videos in varied illumination and weather conditions. 

The recorders labeled the videos informing the scene where some segment were taken (\eg, indoor, urban, crowded environment, \etc), the activity performed (walking,  standing, browsing, driving, biking, eating, cooking, observing, in conversation, \etc), if something caught their attention and when they interacted with some object. Example of labels are depicted in Figure~\ref{fig:dataset}. Also, we create a profile for each recorder representing their preferences over the $80$ classes of the YOLO classifier~\cite{Redmon2016} and the $48$ visual sentiment concepts defined by Sharghi \etal~\cite{Sharghi2017}. To create the recorders' profile, we asked them to indicate their interest in each class and concepts in a scale from $0$ to $10$.

Table~\ref{tab:multimodal_dataset} summarizes in the ``Info'' and ``Videos'' columns the diversity of sensors and activities that can be found in the dataset. Due to the lack of space, we chose the videos which best represent the diverse of activities, camera models, mounting, and the presence/absence of sensors info. The dataset, source code and the 3D model for printing the built setup are publicly available in \href{https://www.verlab.dcc.ufmg.br/semantic-hyperlapse/cvpr2018-dataset/}{https://www.verlab.dcc.ufmg.br/semantic-hyperlapse/cvpr2018}. 

\paragraph{Evaluation criterion.} 
The quantitative analysis presented in this work is based on three aspects: instability, speed-up, and amount of semantic information retained in the fast-forward video.
The \textit{Instability} index is measured by using the cumulative sum over the standard deviation of pixels in a sliding window over the video~\cite{Silva2018}. 
The \textit{Speed-up} metric is given by de difference of the achieved speed-up rate from the required value. The speed-up rate is the ratio between the number of frames in the original video and in its fast-forward version. In this work, we used $10$ as required speed-up. 
For the \textit{Semantic} evaluation, we consider the labels defined in the Semantic Dataset, which characterize the relevant information as pedestrian in the ``Biking'' and ``Driving'' sequences, and face in the ``Walking''. The semantic index is given by the ratio between the sum of the semantic content in each frame of the final video and the maximum possible semantic value (MPSV). The MPSV is the sum of the semantic scores of the $n$ top-ranked frames of the output video, where $n$ is the expected number of frames in the output video, given the required speed-up. 

\subsection{Comparison with state-of-the-art methods}
\begin{table*}[!t]
	\centering
	\caption{Results and videos details of a sample of the proposed multimodal dataset. Duration is the length of the video before the acceleration. RS in the Camera column stands for RealSense\texttrademark~by Intel\textsuperscript{\textregistered} and Hero is a GoPro\textsuperscript{\textregistered} line product.}
	\label{tab:multimodal_dataset}
	\setlength{\tabcolsep}{2.9pt}
	\small{
		\begin{tabular}{lrrrrrrrrrrrrrllcccc} \toprule
			& \multicolumn{2}{c}{\textbf{Semantic}$^1$(\%)} & & \multicolumn{2}{c}{\textbf{Speed-up}$^2$}   & & \multicolumn{2}{c}{\textbf{Instability}$^3$} & & \multicolumn{2}{c}{\textbf{Time}$^3$(s)}     & & \multicolumn{7}{c}{\textbf{Info}} \\  
			\thead{\textbf{Videos}}             & \thead{Ours}          & \thead{MIFF}     & & \thead{Ours}          & \thead{MIFF}   & & \thead{Ours}          & \thead{MIFF}    & & \thead{Ours}          & \thead{MIFF}    & & \thead{Duration \\ (hh:mm:ss)} & \thead{Mount} & \thead{Camera} & & \thead{\rot[90][0.7em]{IMU}} & \thead{\rot[90][0.7em]{Depth}}  & \thead{\rot[90][0.7em]{GPS}}         \\ \cmidrule(l){2-3} \cmidrule(l){5-6} \cmidrule(l){8-9} \cmidrule(l){11-12} \cmidrule(l){14-20} 
			Academic\_Life\_09 & 21.80 & 24.74 & &  0.01 &  0.00 & & 47.56 & 59.38 & &   145.6 & 3,298.5 & & 01:02:53 & helmet   & RS R200      & & \checkmark & \checkmark &            \\
			Academic\_Life\_10 & 24.99 & 25.12 & & -0.02 &  1.53 & & 47.47 & 51.62 & &   282.2 & 7,654.7 & & 02:04:33 & head     & Hero5        & & \checkmark &            & \checkmark \\
			Academic\_Life\_11 & 21.03 & 20.14 & & -0.00 &  0.20 & & 30.19 & 42.64 & &    96.6 & 3,176.9 & & 01:02:04 & hand     & Hero4        & &            &            &            \\
			Attraction\_02     & 65.04 & 59.22 & &  0.10 &  0.00 & & 24.68 & 25.65 & &    95.0 & 5,284.6 & & 01:31:10 & chest    & Hero5        & & \checkmark &            & \checkmark \\
			Attraction\_08     & 80.29 & 77.52 & &  0.35 &  1.72 & & 34.78 & 37.78 & &     8.7 & 1,762.0 & & 00:32:41 & chest    & Hero5        & & \checkmark &            & \checkmark \\
			Attraction\_09     & 43.83 & 44.35 & & -0.18 &  0.29 & & 51.30 & 52.42 & &    27.7 & 3,265.1 & & 00:52:43 & helmet   & RS R200      & & \checkmark & \checkmark &            \\
			Attraction\_11     & 27.28 & 31.55 & & -0.05 & -0.02 & & 31.93 & 35.79 & &   185.6 & 4,779.3 & & 01:17:20 & helmet   & RS R200      & & \checkmark & \checkmark & \checkmark \\
			Daily\_Life\_01    & 18.76 & 20.01 & &  0.04 &  2.56 & & 47.06 & 49.05 & &   126.3 & 5,222.0 & & 01:16:45 & head     & Hero5        & & \checkmark &            & \checkmark \\
			Daily\_Life\_02    & 25.68 & 25.51 & & -0.10 &  3.48 & & 38.16 & 46.80 & &    46.4 & 5,741.3 & & 01:33:39 & head     & Hero5        & & \checkmark &            & \checkmark \\
			Entertainment\_05  & 24.63 & 23.93 & &  0.04 &  0.01 & & 33.79 & 39.12 & &    20.8 & 3,786.1 & & 00:55:25 & helmet   & RS R200      & & \checkmark & \checkmark &            \\
			Recreation\_03     & 76.52 & 72.70 & & -0.04 &  0.45 & & 41.69 & 43.64 & &    37.8 & 3,518.7 & & 00:57:39 & helmet   & Hero4        & &            &            &            \\
			Recreation\_08     & 24.20 & 26.33 & & -0.05 &  3.74 & & 34.98 & 38.44 & &    59.2 & 5,957.0 & & 01:44:15 & shoulder & Hero5        & & \checkmark &            & \checkmark \\
			Recreation\_11     & 67.94 & 65.25 & &  0.20 &  0.02 & & 12.49 & 12.15 & &    17.9 & 2,802.9 & & 00:46:04 & chest    & Hero5        & & \checkmark &            & \checkmark \\
			Sport\_02          & 13.62 & 14.85 & & -0.13 &  6.25 & & 44.96 & 52.59 & &    20.0 & 2,387.6 & & 00:43:20 & head     & Hero5        & & \checkmark &            & \checkmark \\
			Tourism\_01        & 64.00 & 62.90 & & -0.01 &  2.15 & & 28.93 & 31.57 & &    33.6 & 3,283.4 & & 00:55:35 & chest    & Hero4        & &            &            &            \\
			Tourism\_02        & 48.24 & 47.22 & & -0.23 &  3.22 & & 52.38 & 54.27 & &   118.2 & 9,331.0 & & 02:22:52 & head     & Hero5        & & \checkmark &            & \checkmark \\
			Tourism\_04        & 27.20 & 29.24 & &  0.00 &  0.10 & & 53.14 & 56.41 & &   229.4 & 8,302.5 & & 01:46:38 & helmet   & RS R200      & & \checkmark & \checkmark &            \\
			Tourism\_07        & 42.93 & 42.72 & &  0.09 &  4.47 & & 39.44 & 37.08 & &    27.1 & 3,906.1 & & 01:05:03 & head     & Hero5        & & \checkmark &            & \checkmark \\
			\cmidrule(l){2-3} \cmidrule(l){5-6} \cmidrule(l){8-9} \cmidrule(l){11-12}
			\textit{Mean} & \textit{39.89} & \textit{39.63} & &  \textit{0.00} &  \textit{1.72} & & \textit{38.38} & \textit{42.08} & & \textit{87.7} & \textit{4,636.6} & &   &    &   & &  &  &  \\ 
			& \multicolumn{2}{c}{\scriptsize{$^1$\textit{Higher is better.}}} & & \multicolumn{2}{c}{\scriptsize{$^2$\textit{Better closer to 0}}} & & \multicolumn{5}{c}{\scriptsize{$^3$\textit{Lower is better.}}} & & & &  & &  &  & \\ \bottomrule
		\end{tabular}
	}
\end{table*}
In this section, we present the quantitative results of the experimental evaluation of the proposed method. We compare it with the methods: EgoSampling (ES)~\cite{Poleg2015}, Stabilized Semantic Fast-Forward (SSFF)~\cite{Silva2016}, Microsoft Hyperlapse (MSH)~\cite{Joshi2015} the state-of-the-art method in terms of visual smoothness, and Multi-Importance Fast-Forward (MIFF)~\cite{Silva2018} the state-of-the-art method in terms of the amount of semantics retained in the final video.

% Semantic
Figure~\ref{fig:results_baselines}-a shows the results of the Semantic evaluation performed using the sequences in the Semantic Dataset. We use the area under the curves as a measure of the retained semantic content. Our  approach outperformed the other methodologies. The area under the curve of the proposed method was $100.74\%$ of the area under the MIFF curve, which is the state-of-the-art in semantic hyperlapse. The second Semantic Hyperlapse technique evaluated, SSFF, had $52.01\%$ of the area under curve of MIFF. Non-semantic hyperlapse techniques such as MSH and ES achieved at best $19.63\%$ of the MIFF area.

% Instability
The results for Instability are presented as the mean of the instability indexes calculated over all sequences in the Semantic Dataset (Figure~\ref{fig:results_baselines}-b, lower values are better). The black dotted and the green dashed lines stand for the mean instability index when using an uniform sampling and for the original video, respectively. Ideally, it is better to yield an instability index as closer as possible to the original video. The reader is referred to the Supplementary Material for the individual values. The chart shows that the our method created videos as smooth as the state-of-the-art method MSH and smoother than the MIFF. 
%This good performance is due to the Frame and Speed-up Transition Smoothing, as it is discussed in Section~\ref{subsec:ablation}.

% Speed-up
Figure~\ref{fig:results_baselines}-c shows the speed-up achieved by each method. The bar represent the average difference between the required speed-up and the rate achieved by a respective method for each class of video in the Semantic Dataset. Values closer to zero are desirable. The chart shows that our method provided the best acceleration for ``Driving'' and ``Walking'' experiments. In ``Biking'' experiments MSH held the best speed-up.

As far as the semantic metric is concerned \mbox{(Figure~\ref{fig:results_baselines}-a)}, our approach leads followed by MIFF. We ran a more detailed performance assessment comparing our method to MIFF in the multimodal dataset. The results are shown in Table~\ref{tab:multimodal_dataset}. As can be seen, our method outperforms MIFF in all metrics. The column ``Time'' shows the time for the frame sampling step of each method (MIFF runs a parameter setup and the shortest path, and ours runs minimum reconstruction followed by the smoothing step). Our method was ${53\times}$ faster than MIFF. It is noteworthy that, unlike MIFF that requires $14$ parameters to be adjusted, our method is parameter-free. Therefore, the average processing time spent per frame was $0.5$~ms, while the automatic parameter setup process and the sampling processing of MIFF spent $30$~ms per frame. The descriptor extraction for each frame ran in $320$~ms facing $1{,}170$~ms of MIFF. The experiments were conducted in a machine with i7-6700K CPU @ 4.00GHz and 16 GB of memory.

\subsection{Ablation analysis}
\label{subsec:ablation}

In this Section, we discuss the gain of applying the steps Weighted Sparse Frame Sampling and Smoothing Frame Transitions to the final fast-forward video. All analysis were conducted in the Semantic Dataset.

\paragraph{Weighted Sparse Sampling.}
As stated, we introduce a new model based on weighted sparse sampling to address the problem of abrupt camera motions. In this model, small weights are applied to frames containing abrupt camera motions to increase the probability of these frames being selected and, consequently, to create a smooth sequence.

\begin{figure}[!t]
	\centering
	\includegraphics[width=1.0\linewidth]{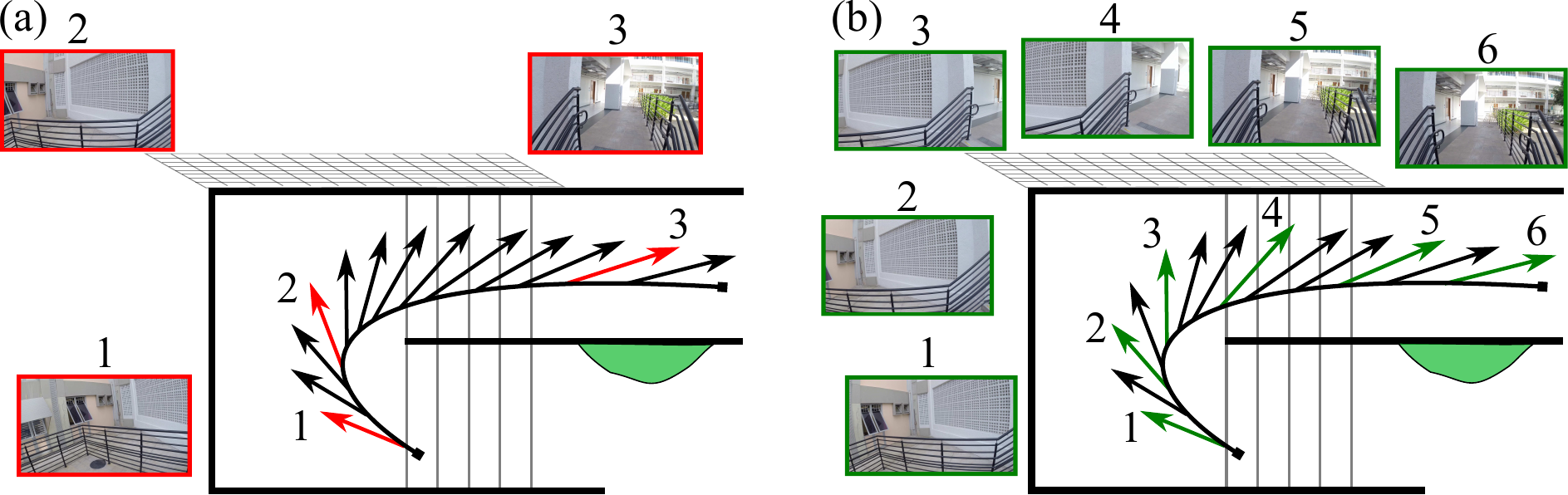}
	\caption{The effect of applying the Weighted Sparse Sampling in an abrupt camera movement segment. Black arrows are the frames of the original video, red arrow are frames selected by non-weighted sparse sampling, and the green arrows represent the frames sampled by the weighted sparse sampling. Each image is related with the respective numerated arrow.}
	\label{fig:results_weighted}
\end{figure}

\begin{figure}[!t]
	\centering
	\includegraphics[width=1.0\linewidth]{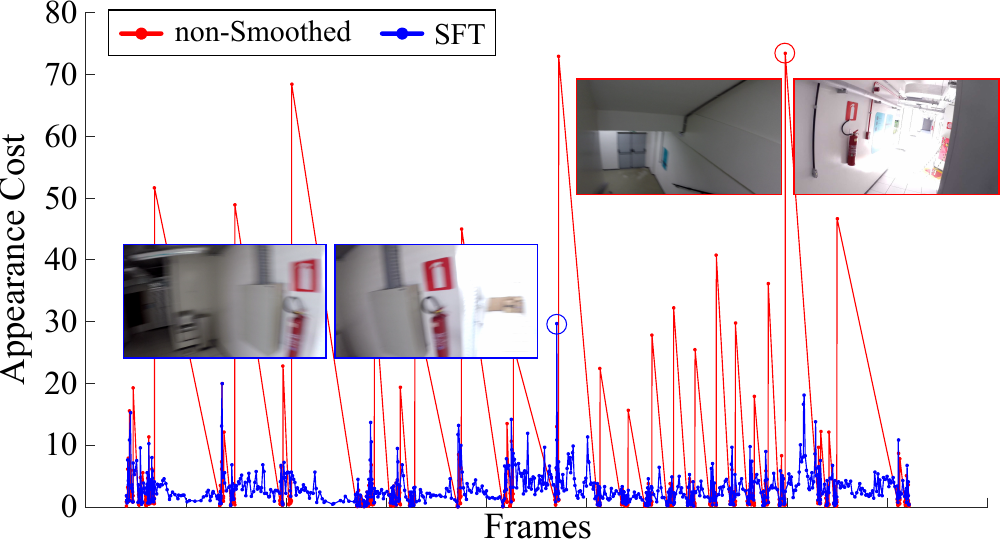}
	\caption{Frame sampling and appearance cost of the transitions in the final video before and after applying the Smoothing Frame Transition (SFT) to the video ``Walking 25p''. Images with blue border show the frames composing the transition with the highest appearance cost using SFT. Images with red borders are related to the non-smoothed sparse sampling.}
	\label{fig:results_SFT}
\end{figure}

Considering all sequences of abrupt camera motions present in all videos of the Semantic Dataset, the weighted version manages to sample, in average, three times more frames than the non-weighted version. Figure~\ref{fig:results_weighted} illustrates the effect of  solving the sparse sampling by weighting the activation vector. It can be seen that the weighting strategy helps by using a denser sampling in curves (on the right) than when applying the non-weighted sparse sampling version (on the left). In this particular segment, our approach sampled twice the number of frames, leading to less shaky lateral motions. 
\paragraph{Smoothing Frame Transitions.}
By computing the coefficient of variation (CV), we measured the relative variability of the points representing the appearance cost of the frames (blue and red points in Figure~\ref{fig:results_SFT}). The appearance cost is computed as the Earth Mover's Distance~\cite{Pele2009} between the color histogram of frames in a transition. 

After applying the proposed smoothing approach we achieved ${CV=0.97}$, while the simple sampling provided ${CV=2.39}$. The smaller value for our method indicates a smaller dispersion and consequently fewer visual discontinuities. Figure~\ref{fig:results_SFT} shows the result when using SFT and non-smoothed sparse sampling. The horizontal axis contains the index of selected frames and the vertical axis represents the appearance cost between the $i$-th frame and its following in the final video. The points in the red line represent the oversampling pattern of non-smoothed sparse sampling, in which many frames are sampled in segments hard to reconstruct followed by a big jump. 

The abrupt scene changing is depicted by high values of appearance cost. The red-bordered frames in the figure show an example of two images that compose the transition with the highest appearance cost for a fast-forwarded version of the video ``Walking 25p'' using non-smoothed sparse sampling. After applying the SFT approach, we have a more spread sampling covering all segments, and with less video discontinuities. The blue-bordered images present the frames composing the transition with the highest appearance cost using the sparse sampling with the SFT step. By comparing the red and blue curves, one can clearly see that after using SFT, we achieve smoother transitions, \ie, lower values for the appearance cost.
\section{Conclusion}
\label{sec:conclusion}

In this work, we presented a new semantic fast-forward parameter-free method for first-person videos. It is based on a weighted sparse coding modeling to address the adaptive frame sampling problem and smoothing frame transitions to tackle abrupt camera movements by using a denser sampling along the segments with high movement. Contrasting with previous fast-forward techniques that are not scalable in the number of features used to describe the frame/transition, our method is not limited by the size of feature vectors.

The experiments showed that our method was superior to state-of-the-art semantic fast-forward methods in terms of semantic, speed-up, stability, and processing time. We also performed an ablation analysis that showed the improvements provided by the weighted modeling and smoothing step. An additional contribution of this work is a new labeled $80$-hour multimodal dataset, with several annotations related to the recorder preferences, activity, interaction, attention, and the scene where the video was taken.

\paragraph{Acknowledgments.}
The authors would like to thank the agencies CAPES, CNPq, FAPEMIG, and Petrobras for funding different parts of this work.

{\small
\bibliographystyle{ieee}
\bibliography{ref_CVPR2018}
}

\end{document}